\documentclass[letterpaper, 10 pt, conference]{ieeeconf}
\IEEEoverridecommandlockouts
\usepackage{graphics} % for pdf, bitmapped graphics files
\usepackage{wrapfig} % for dense pic, with text side by side
\usepackage[linesnumbered,ruled]{algorithm2e}
\graphicspath{{img/}}
\usepackage{epsfig} % for postscript graphics files
\usepackage{amsmath} % assumes amsmath package installed
\usepackage{amssymb}  % assumes amsmath package installed
\usepackage{amsthm}
\usepackage{bigints}
\usepackage{pifont}
\def\-{\raisebox{.75pt}{-}}
\usepackage{mathtools}
\usepackage{cite}
\usepackage[makeroom]{cancel}
\usepackage[table]{xcolor} 
\usepackage[center]{subfigure}
\usepackage{multirow}
\usepackage{hyperref}
\usepackage{booktabs}
\usepackage{subfigure}
\usepackage{epstopdf}
\usepackage{textcomp}
\usepackage{makecell}
\usepackage{bm}
\usepackage{caption}
\usepackage{algpseudocode,algorithm} % Algorithm environment
\usepackage{color, soul}
\usepackage{mathptmx}
\usepackage[11pt]{moresize}
\usepackage{hhline}
\definecolor{ForestGreen}{RGB}{34,139,34}

\usepackage{tikz}
\usepackage{textcomp}
\def\BibTeX{{\rm B\kern-.05em{\sc i\kern-.025em b}\kern-.08em
    T\kern-.1667em\lower.7ex\hbox{E}\kern-.125emX}}

\usepackage{eso-pic}

\newcommand{\subparagraph}{}
\definecolor{amaranth}{rgb}{0.9, 0.17, 0.31}
\definecolor{bleudefrance}{rgb}{0.19, 0.55, 0.91}
\begin{document}
\title{Priors are Powerful: Improving a Transformer for \\ Multi-camera 3D Detection with 2D Priors \thanks{$^*$Work done while both authors were at Argo AI (\url{https://www.argo.ai/}). Di Feng is now with Apple, and Francesco Ferrorni is with Nvidia. Correspondence: \url{fengdi1015@gmail.com}}
}
\author{Di Feng$^{*}$, Francesco Ferroni$^{*}$}
\maketitle

\begin{abstract}
Transfomer-based approaches advance the recent development of multi-camera 3D detection both in academia and industry. In a vanilla transformer architecture, queries are randomly initialised and optimised for the whole dataset, without considering the differences among input frames. In this work, we propose to leverage the predictions from an image backbone, which is often highly optimised for 2D tasks, as priors to the transformer part of a 3D detection network. The method works by (1). augmenting image feature maps with 2D priors, (2). sampling query locations via ray-casting along 2D box centroids, as well as (3). initialising query features with object-level image features. Experimental results shows that 2D priors not only help the model converge faster, but also largely improve the baseline approach by up to $12\%$ in terms of average precision. 
\end{abstract}

%\copyrightnotice\copyrightnotice
%\setlength{\parskip}{2mm plus3mm minus3mm}\setlength{\belowdisplayskip}{8pt} %\setlength{\belowdisplayshortskip}{8pt}
%\setlength{\abovedisplayskip}{8pt} 
%\setlength{\abovedisplayshortskip}{8pt}

%-------------------------------------------------------------------------
\section{Introduction} \label{sec:introduction}
%-------------------------------------------------------------------------
% multi-camera 3d detection problem
Towards 360-degree 3D perception, self-driving vehicles are usually equipped with multiple monocular cameras, and reliable and accurate multi-camera 3D detection has become an important research challenge and industrial effort. The traditional approach takes advantage of the convolution neural nets (convnets) that are highly-optimised for 2D tasks, by performing 2D scene understanding on images, followed by a 2D to 3D projection. Recent advancements, in contrast, propose to project 2D images into 3D space, before running 3D tasks on a bird's eye view (BEV) representation~\cite{tesla}. This new paradigm not only provides a generic scene representation for multi-modal perception, mapping, and prediction, but also achieves improved accuracy with the help of the \textit{transformer} architecture~\cite{li2022delving}. 

\begin{figure}[!tpb]
\centering
\begin{minipage}{1\linewidth}
	\centering
\includegraphics[width=1.0\textwidth]{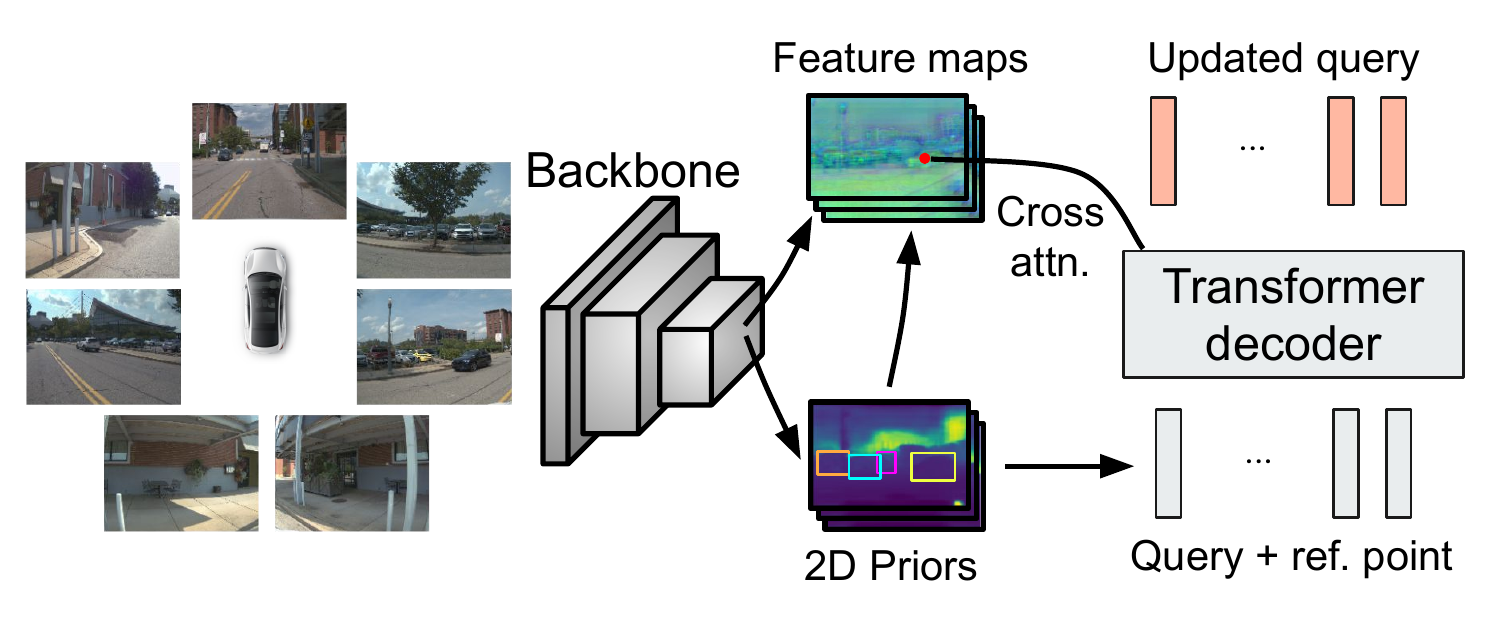}
	\caption{The pipeline for multi-camera 3D detector with transformers. First, a backbone network, commonly a convolution neural network (convnet), extracts multi-level feature maps from multi-camera inputs. Afterwards, a transformer decoder iteratively processes queries and interacts with feature maps. Finally, each query is fed into a detection head to predict object classes and cuboid parameters. In this work, we propose to leverage 2D predictions from the convnet backbone as priors to the transformer decoder for 3D detection. Those priors are incorporated into feature maps, as well as each query and reference point.}
 \label{fig:architecture}
\end{minipage}
\end{figure}

\begin{figure*}[htpb]
\centering
\begin{minipage}{1\linewidth}
	\centering
	\subfigure[Vanilla detr3d]{\label{fig:method_detr3d}\includegraphics[width=0.49\textwidth]{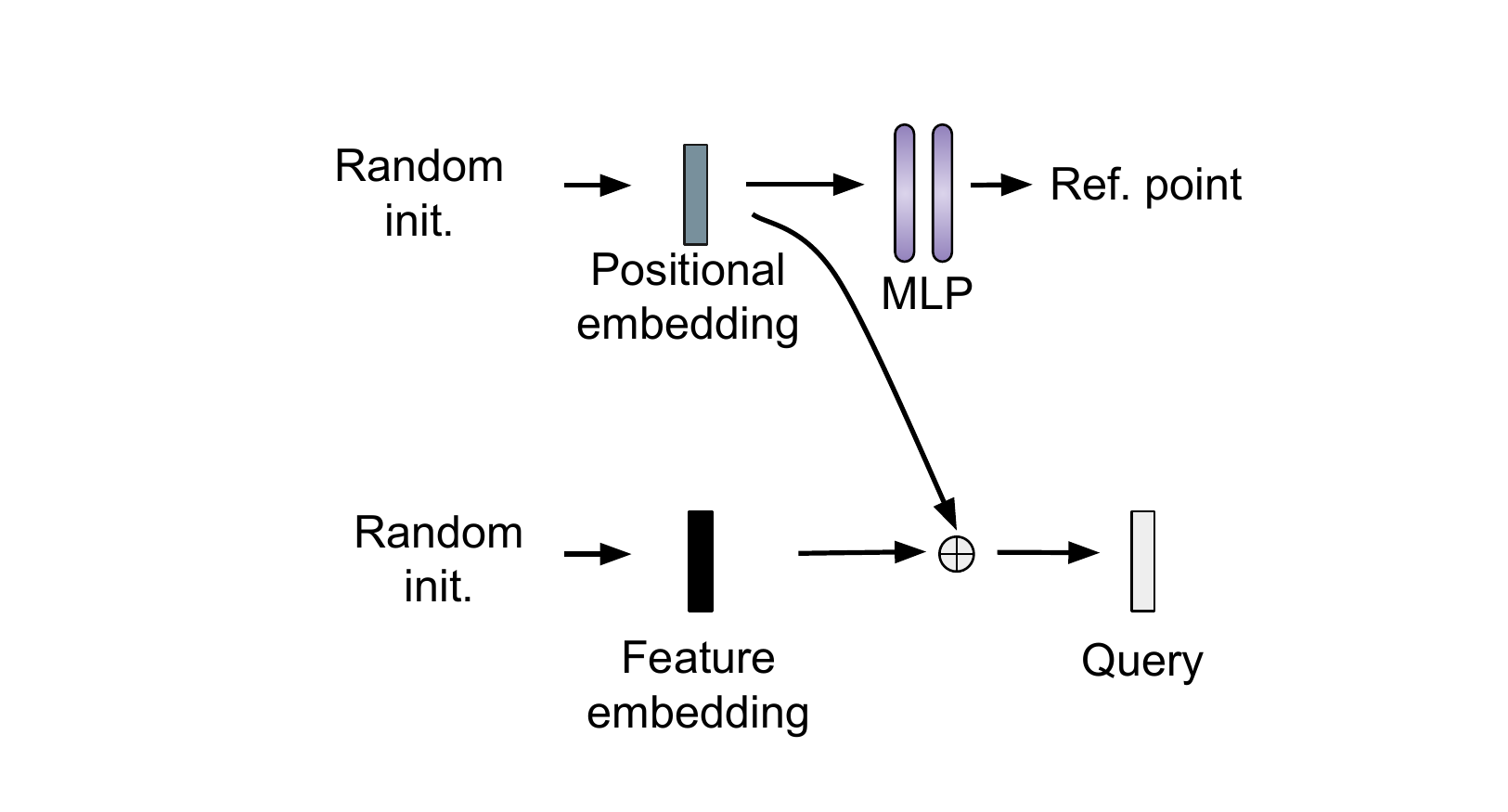}}
    	\subfigure[Ours]{\label{fig:method_ours}\includegraphics[width=0.49\textwidth]{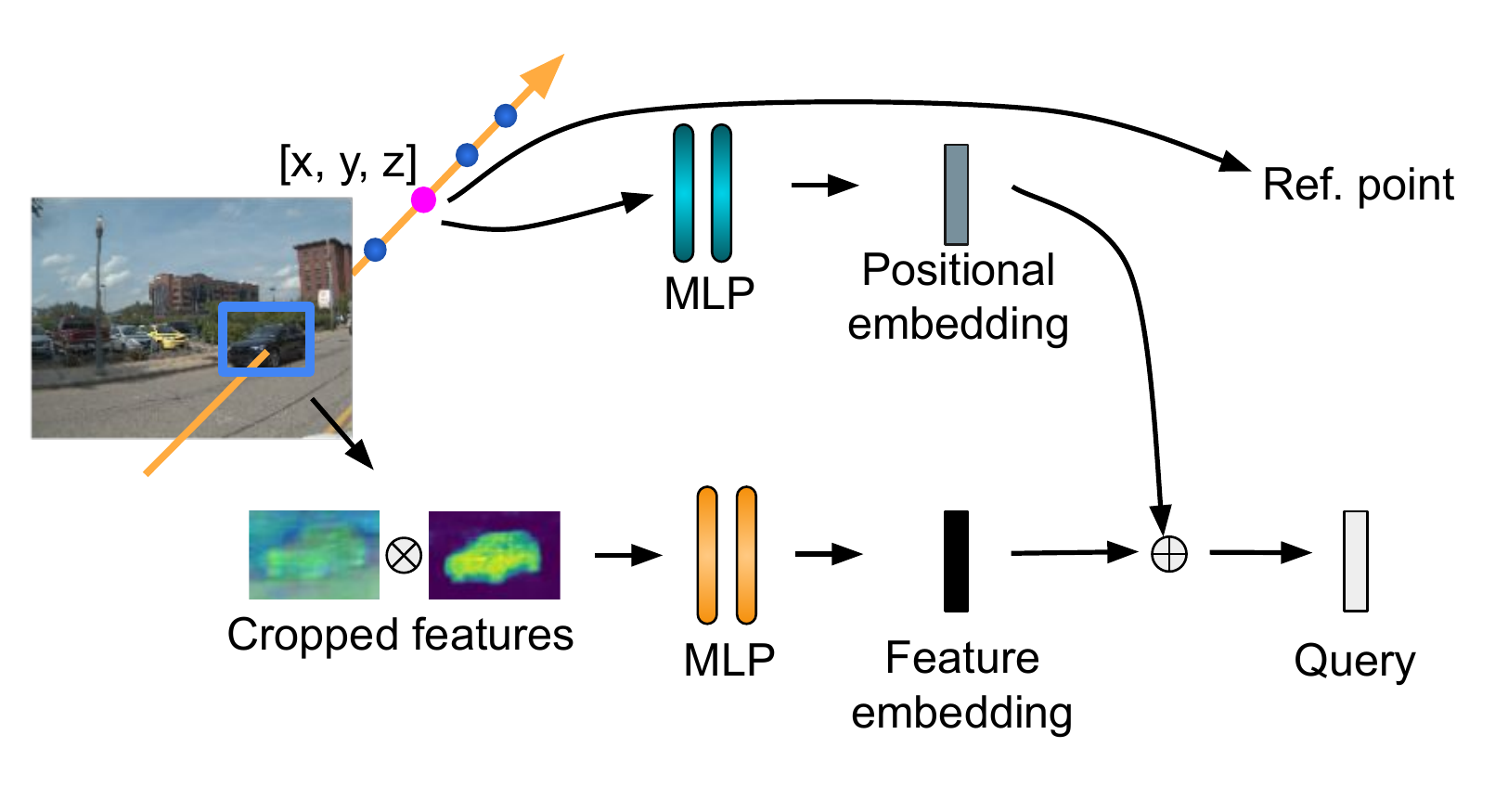}}
	\caption{A comparison of query generation strategies between the vanilla detr3d~\cite{wang2022detr3d} and our proposed methods with 2D priors. (a). The vanilla detr3d randomly initialises the positional embedding and the feature embedding vectors. Reference points are predicted by a small multi-layer perceptron (MLP). (b). Our proposed approach leverages 2D detection and semantic maps as 2D priors. We sample reference points via ray-casting along 2D box centroids, which generates the positional embedding vector through a MLP. Besides, we initialize the feature embedding vector with the object-level features, weighted by semantic scores.}
\end{minipage}
\end{figure*}

A typical pipeline for multi-camera 3D detection with transformers is shown in Fig.~\ref{fig:architecture}. First, multi-level feature maps from multiple camera images are extracted from a backbone network, commonly a convnet. Afterwards, a transformer decoder iteratively processes queries and interacts with image feature maps via cross-attention~\cite{carion2020end}. Finally, each updated query is fed into a detection head to categorize objects and regress their cuboid parameters (such as centroid locations, cuboid extents, and orientations). In a vanilla transformer architecture, such as from the seminal work detr3d~\cite{wang2022detr3d}, query features and their location information are randomly initialised and optimised for the whole dataset, without considering the heterogeneity of inputs from different frames. We find that such query design suffers from slow training convergence and strong smearing effect with erroneous depth estimation (shown in Fig.~\ref{fig:method_detr3d} and will be discussed in Sec.~\ref{sec:vanilla}). 

\begin{figure}[htpb]
\centering
\begin{minipage}{1\linewidth}
	\centering
	\subfigure[Predictions from detr3d]{\label{fig:smearing}\includegraphics[width=0.49\textwidth]{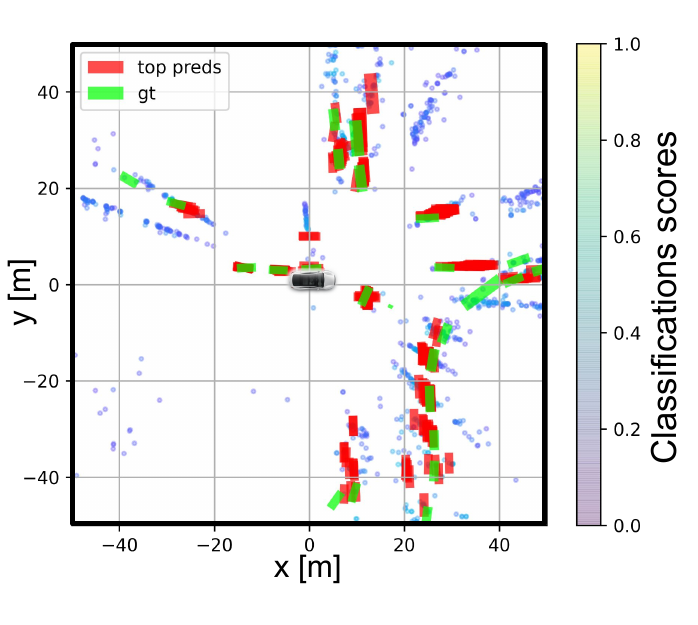}}
    	\subfigure[Query sampling results]{\label{fig:loc_priors}\includegraphics[width=0.49\textwidth]{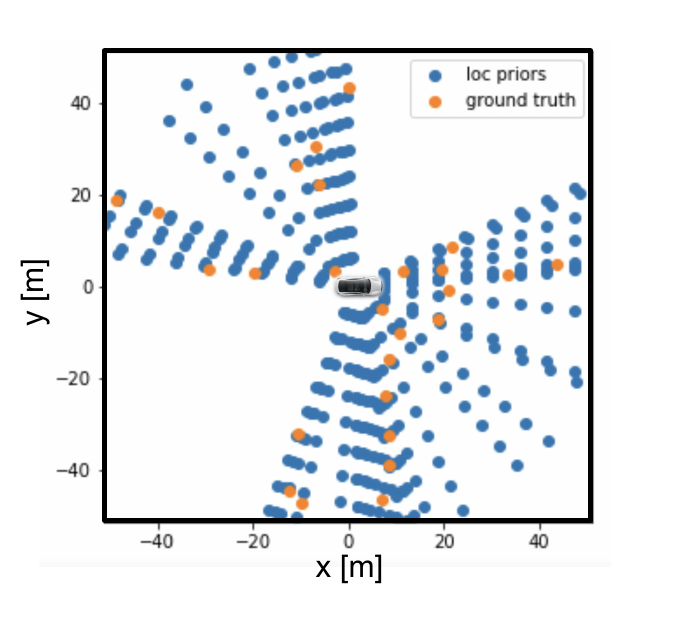}}
	\caption{(a). Raw predictions from the vanilla detr3d model on the Bird's Eye View (BEV). Each circle represents a query prediction. Strong smearing effect can be observed along the ray. (b). Our proposed query sampling strategy. Blue dots are generated reference points, and orange dots are ground truth centroids. All ground truth centroids can be associated with nearby reference points, though with some errors.}
\end{minipage}
\end{figure}

Several methods extend detr3d~\cite{wang2022detr3d} with improved query design. For example, PETR~\cite{liu2022petr} generates 3D position embedding to image features as the input of a transformer decoder.  SpatialDETR~\cite{Doll2022ECCV} encodes camera intrinsics and extrinsics features to keys and queries. Graph-detr3d~\cite{chen2022graph} replaces self-attention with a graph neural network for better query interaction. Finally, BEVFormer~\cite{li2022bevformer} discretises the 3D world with bird's eye view grids, and considers each grid as a query location. 
 
Since convnets are often highly optimised for 2D tasks, why not reusing those 2D predictions as priors to the transformer part of 3D detection? In this work, we verify this idea in the detr3d pipeline, and incorporate 2D object detection, semantic segmentation, and depth estimation from a image backbone to the transformer decoder. We propose three simple strategies to use 2D priors: augmenting image feature maps for cross attention, sampling query locations via ray-casting along 2D box centroids, as well as initialising query features with object-level image features. Experimental results on an internal dataset shows that our methods largely improve the vanilla detr3d by up to $12\%$ in terms of average precision, and make the model converge faster during training. 

In parallel to our work, MV2D~\cite{wang2023object} also proposes to leverage 2D detections as priors for the transformer part of a multi-camera 3D detector. Unlike our approaches which generate multiple reference points from a 2D box centroid and employs multiple 2D cues (2D boxes, semantic maps, and depth maps), MV2D only studies how to exploit 2D detections, and how to predict one reference point for each 2D box via a dynamic object query generator. Their experimental results demonstrate higher recall rates compared to the vanilla transformer model, especially for small and distant objects.

In the sequel, Sec.~\ref{sec:vanilla} reviews the transformer decoder part of the detr3d model, with a focus on the query generation process. Sec.~\ref{sec:methods} introduces our proposed three methods to improve the detr3d network. Sec.~\ref{sec:results} shows the experimental results, followed by a summary and discussion in Sec.~\ref{sec:summary}.

%-------------------------------------------------------------------------
\section{Vanilla Detr3d Revisited} \label{sec:vanilla}
%-------------------------------------------------------------------------
The architecture of detr3d~\cite{wang2022detr3d} has been summarised in the previous section and depicted in Fig.~\ref{fig:architecture}. In this section, let us take a closer look at the transformer decoder. 

A decoder is built from six standard transformer blocks~\cite{carion2020end}. In each block, queries are interacted with each other via self-attention, and fused with multi-camera multi-level feature maps via cross-attention. Unlike the common ``global" cross-attention mechanism~\cite{carion2020end}, detr3d only associates a query to image features that correspond to its query location (also called reference point). To do this, a position $p=[x,y,z]$ in the 3D coordinate system is computed for each query. The position is projected onto image planes given camera intrinsics and extrinsics parameters. The image features from the projected pixels are weighted and averaged over all feature levels and cameras for updating a query in a ``local" cross-attention manner. 

Denote $d$ as the embedding dimension for a query, Fig.~\ref{fig:method_detr3d} illustrates how a query $q\in \mathbb{R}^d$ and its reference point $p\in \mathbb{R}^3$ are built. First, a position embedding vector $q_{\text{pos}}\in \mathbb{R}^d$ and a feature embedding vector $q_{\text{feat}}\in \mathbb{R}^d$ are randomly initialized (following a uniform or a normal distribution). Afterwards, $q_{\text{pos}}$ is mapped to the reference point $p$ via a multi-layer perceptron (MLP), and added to $q_{\text{feat}}$ to generate the final positional-aware query features $q$. Through training the network with the standard Hungarian assignment and the set prediction loss~\cite{carion2020end}, both $q_{\text{pos}}$ and $q_{\text{feat}}$ learn to encode the object statistics for the whole dataset, which can be considered as pre-defined ``anchors" in common object detection pipeline, such as Faster-RCNN~\cite{girshick2015fast}. 

Though simple and straightforward, the network design in~\cite{wang2022detr3d} lacks prior knowledge in the query and reference point generation process, resulting in slow training convergence and ambiguity in prediction. Fig.~\ref{fig:smearing} illustrates a typical detr3d output on the bird's eye view (BEV) without post-processing. All queries are marked by circles, and those with high classification scores are further demonstrated with red bounding boxes. We observe strong smearing effect in object detection, i.e. queries are converged along the ray from detections on 2D images, bringing many false positives. This is due to erroneous depth estimation and ambiguous target assignment during training\footnote{Imagine multiple queries generate reference points, which are close to each other and are projected to the same object on images. Due to the Hungarian assignment~\cite{carion2020end}, only one query is labelled ``positive", punishing other positive queries with ``negative" signals.}.

%-------------------------------------------------------------------------
\section{Three Ways of Adding Priors} \label{sec:methods}
%-------------------------------------------------------------------------
We propose three methods to improve the detr3d network, by incorporating 2D priors to the transformer decoder, illustrated in Fig.~\ref{fig:method_ours}. To do this, we select 2D object detection, semantic segmentation, and depth estimation predicted by our convnet backbone as priors, as they are common, well-optimized 2D tasks for autonomous driving (e.g. HydraNet from Tesla~\cite{tesla}). The depth estimation is represented as a single-channel depth map rescaled to $[0, 1]$. The semantic segmentation is represented by a semantic map with $C$ channels, where each channel shows pixel-wise classification scores of a category.

\subsection{Feature Map Priors}
We simply concatenate semantic and depth maps with the multi-camera feature maps at different scales. In this way, semantic and depth priors are added to queries in a cross-attention operation.

\subsection{Location Priors}
We generate reference points $p\in \mathbb{R}^3$ only along the rays from the centroids of 2D box predictions. For each ray, a simple uniform sampling with $5$ meters interval is performed. In this way, the search space for objects can be narrowed efficiently, which helps reduce false positives, limit the number of queries, and accelerate model convergence. When performing cuboid prediction, the detection head regresses the offset to its reference point, denoted by $\Delta x, \Delta y, \Delta z$, as the cuboid centroids. It also regresses the cuboid's height, length, width, and yaw angle. 

The reference points may not accurately overlap with the cuboid centroids, because the centers of 2D boxes are different from those from the projected cuboids, and the $5$-meters sampling interval is rough compared to the common discretisation thresholds in many well-known detection networks (e.g 0.5-meters interval in Lift-spalt-shoot~\cite{philion2020lift}, 0.16-meters in CaDNN~\cite{reading2021categorical}, and 0.2-meters in Pointpillar~\cite{lang2019pointpillars}). However, we find such a simple point generation strategy provides rough-and-ready estimates to the actual object locations, as illustrated in Fig. \ref{fig:loc_priors}. Besides, the location errors can be compensated by the iterative query refinement in transformer blocks. We expect more accurate reference point generation, when introducing a centerness head for projected cuboid centers (similar to CenterNet~\cite{zhou2019objects}), or sampling points only around predicted depth (similar to CramNet~\cite{hwang2022cramnet}).

Inspired by Anchor-DETR~\cite{wang2022anchor}, we further incorporate location priors to queries, by projecting a reference point $p\in \mathbb{R}^3$ to a position embedding vector $q_{\text{pos}}\in \mathbb{R}^d$ via a small MLP. Interestingly, this is a reversed procedure compared to the vanilla detr3d, which maps a position embedding vector to its reference point. 

\subsection{Query Priors}
All queries generated from a ray come from the same 2D object. Therefore, we propose to incorporate the same object-level 2D priors to those queries, and further distinguish among them with positional information. We follow five steps: First, the semantic map, the depth map, and the multi-level multi-camera feature maps are cropped based on the 2D box estimates. Then, the channel of the cropped semantic map, which corresponds to the predicted object class, is used to weight the cropped depth map and feature maps by a pixel-wise dot production. Afterwards, a channel-wise global average-pooling operation is used to generate a 1D vector for each query prior, inspired from the squeeze-and-excitation operation from SENet \cite{hu2018squeeze}. Furthermore, the query prior vector, appended with an object class index, an objectness score, and the 2D bounding box parameters, is fed into a small MLP to generate query embedding features. Finally, the positional embedding features are added to the query embedding features as the final query features, so that the queries from the same ray are distinguishable.

%-------------------------------------------------------------------------
\section{Experimental Results} \label{sec:results}
%-------------------------------------------------------------------------
\begin{figure*}[htpb]
\centering
\begin{minipage}{1\linewidth}
	\centering
	\subfigure[]{\includegraphics[width=0.32\textwidth]{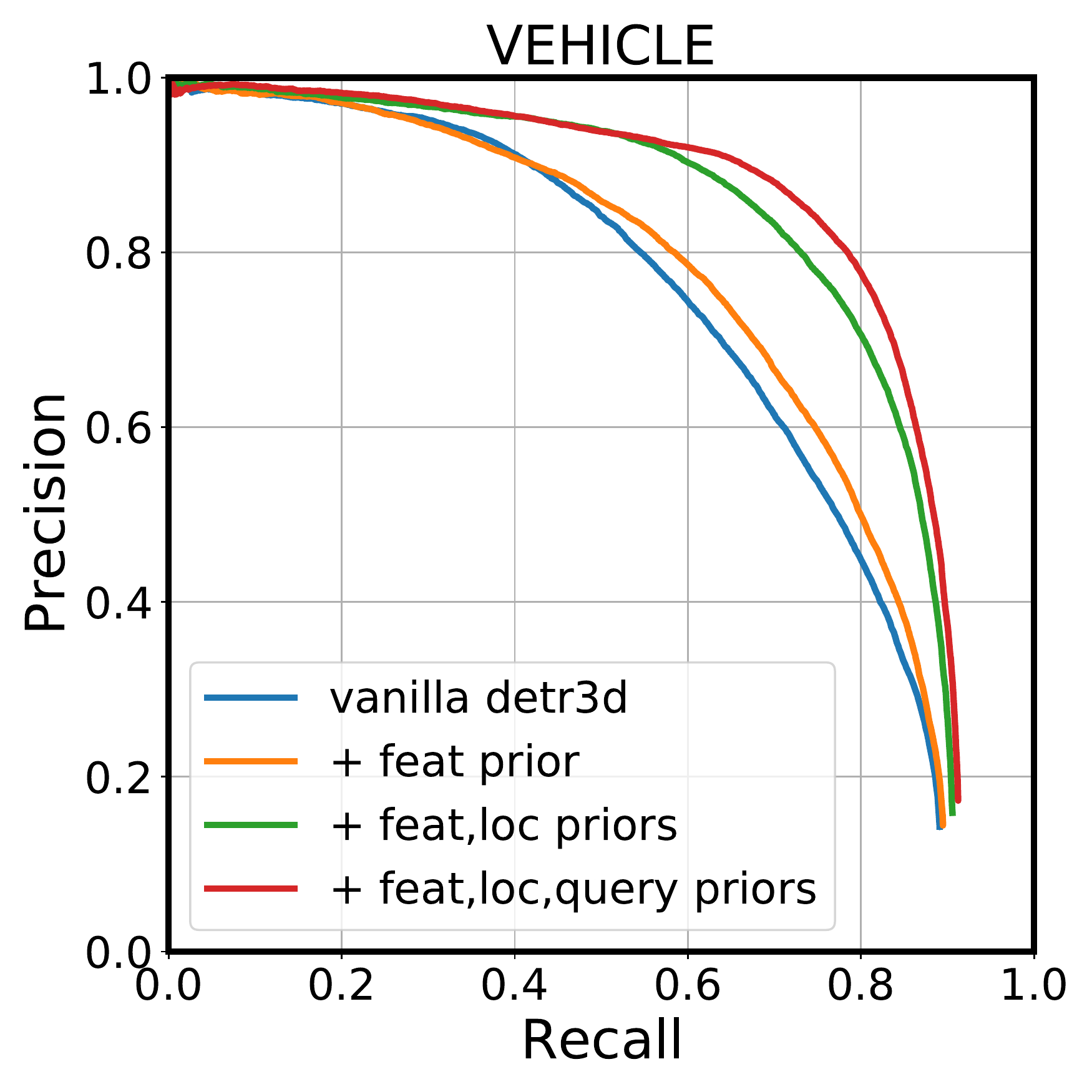}\label{fig:pr_curve_vehicle}}
    	\subfigure[]{\includegraphics[width=0.32\textwidth]{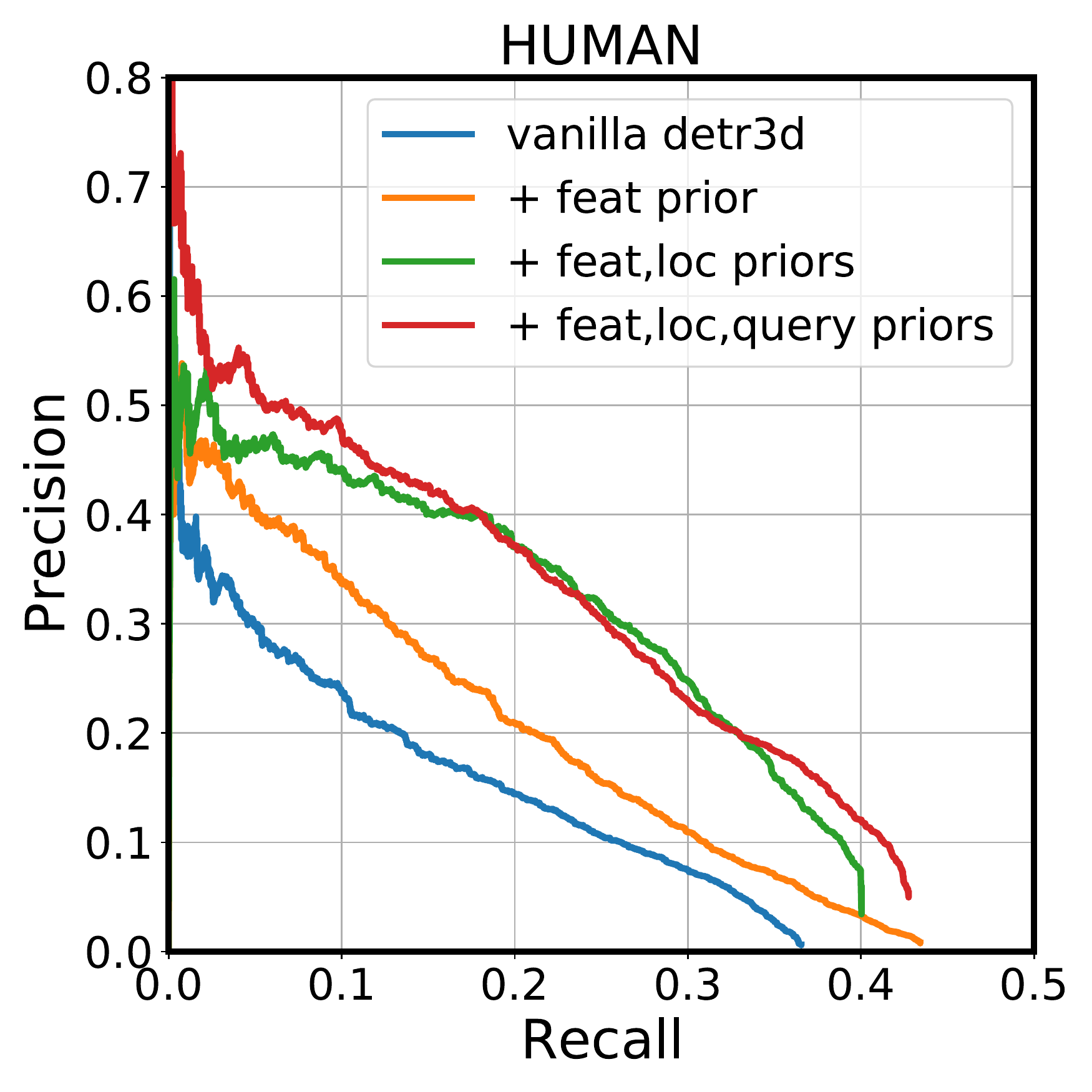}\label{fig:pr_curve_human}}
	\subfigure[]{\includegraphics[width=0.32\textwidth]{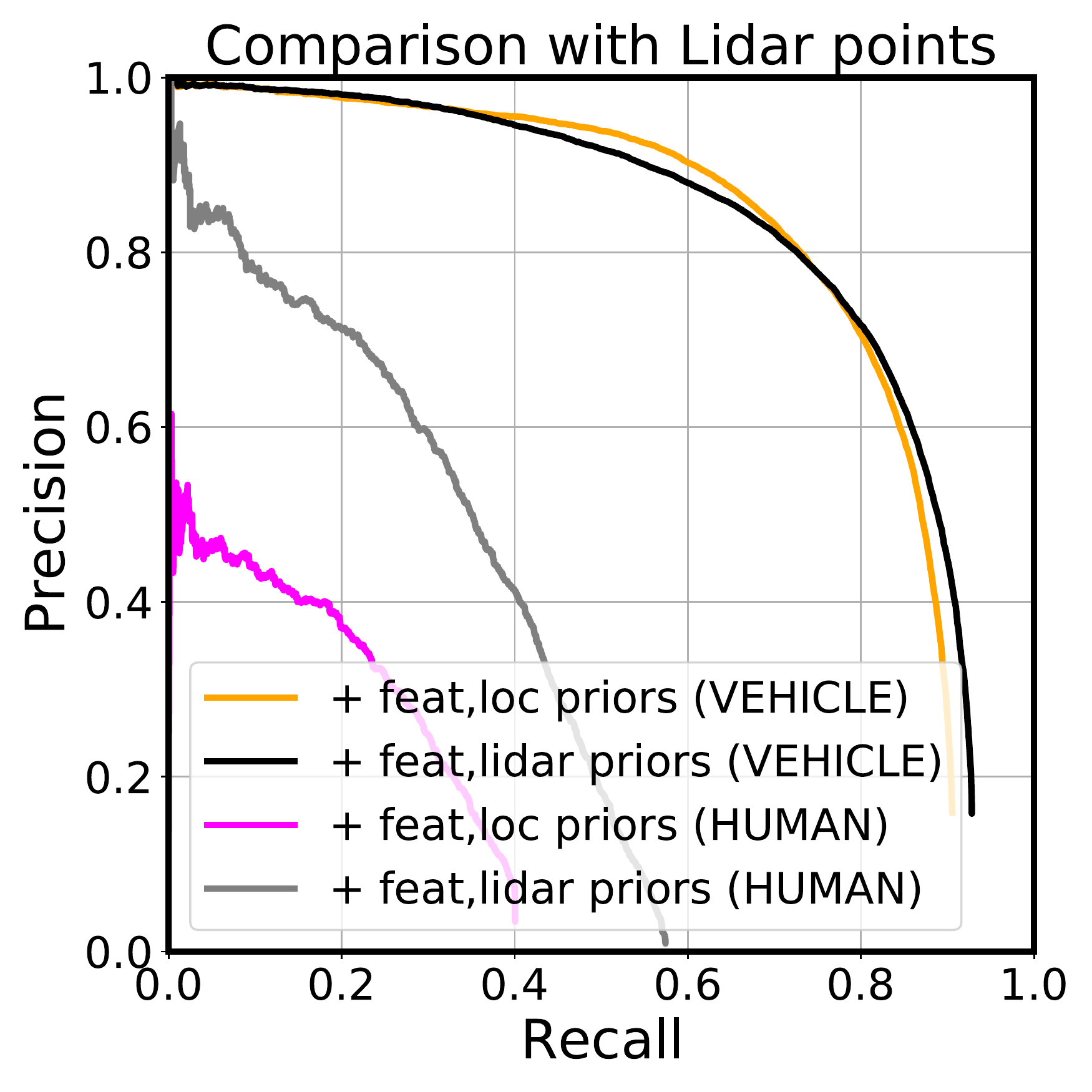}\label{fig:pr_curve_lidars}}
	\caption{Precision recall curves. (a). A comparison among the vanilla detr3d model and its variants with 2D priors on the VEHICLE class. (b). A comparison on the HUMAN class. (c). Ablation study by replacing reference points with lidar observations.}
\end{minipage}
\end{figure*}

Based on a pre-trained convnet backbone, we re-implement the detr3d transformer decoder, and experiment its detection performance with different 2D priors. Following the original detr3d~\cite{wang2022detr3d}, we set the initial learning rate to be $2*10^{-4}$ with a weight decay of $10^{-5}$. The AdamW optimiser with a consine decay is used. Unlike \cite{wang2022detr3d}, we do not use any data augmentation tricks, and find that training with more epochs improve the model performance. All models, unless mentioned otherwise, are trained with a tiny subset of our internal dataset, with approx. 60k training, 10k validation, and 4k testing samples. The data was recorded in various locations in the US and Europe, with different lightning conditions (daytime, nighttime, rainy, sunny etc.) and scenarios (cities, rural areas, etc.). We report the Average Precision (AP) scores at the IoU=$0.1$ threshold on the bird's eye view (BEV) for the VEHICLE and HUMAN classes, and only consider detections within the $50$ meters range.

\subsection{Main Results}\label{subsec:main_results}
Tab.~\ref{tab:comparison} compares the AP scores between the vanilla detr3d model with its variants with different 2D priors. The model ``+ feat prior " only adds feature map priors, ``+ feat, loc priors" additionally uses location priors, and ``+ feat, loc, query priors" exploits all three priors. Fig.~\ref{fig:pr_curve_vehicle} and Fig.~\ref{fig:pr_curve_human} show the precision recall curves for VEHICLE and HUMAN classes, respectively. We observe that all 2D priors improve the vanilla detr3d model with higher AP scores up to nearly $12\%$. The largest performance gain comes from location priors, verifying the effectiveness of our design choice for reference point generation.
\begin{table}[htpb]
\begin{center}
\resizebox{1.0\linewidth}{!}{
\begin{tabular}{|l|c|c|}
\hline
Models & AP  VEHICLE ($\%$) & AP  HUMAN ($\%$) \\
\hline\hline
vanilla detr3d & 70.57 & 6.36 \\ \hline
+ feat prior & 72.40 (+1.83) & 9.16 (+1.80) \\ 
+ feat, loc priors & 79.93 (+9.36) & 13.52  (+7.16)\\ 
+ feat, loc, query priors & 82.01 (+11.44) & 14.68 (+8.32) \\
\hline
\end{tabular}}
\end{center}
\caption{A comparison of Average Precision (AP) scores at the IoU=$0.1$ threshold.}\label{tab:comparison}
\end{table}
\begin{table}[htpb]
\begin{center}
\resizebox{1.0\linewidth}{!}{
\begin{tabular}{|l|c|c|}
\hline
Models & AP  VEHICLE ($\%$) & AP  HUMAN ($\%$) \\
\hline\hline
vanilla detr3d & 75.84 & 24.79 \\ 
+ feat, loc, query priors & 86.52 (+10.68) & 58.06 (+33.27) \\
\hline
\end{tabular}}
\end{center}
\caption{A comparison of Average Precision (AP) scores at 4 meters centroid distance threshold.}\label{tab:comparison_4m}
\end{table}

In addition, Tab.~\ref{tab:comparison_4m} reports AP scores at the 4 meters threshold, which are commonly used in the Nuscenes metrics~\cite{caesar2020nuscenes}. This threshold is less strict than the IoU=$0.1$ threshold when evaluating location errors, thus resulting in higher AP scores when evaluating on the same model. In this setting, we observe that the model with 2D priors largely improves the HUMAN detection by more than $30\%$.    

\subsection{Using Lidar Points as Location Priors}
We conduct a simple ablation study by replacing the location priors with Lidar point clouds. To do that, we train a model called ``+ feat, lidar priors", which uses the uniformly sub-sampled lidar observations as reference points. Fig.~\ref{fig:pr_curve_lidars} shows that our camera-only model ``+ feat, loc priors" achieves similar performance with its camera-lidar fusion counterpart when detecting the VEHICLE class, but performs much worse for the HUMAN class. The result indicates that localization errors are still the bottleneck for the camera-only detection pipeline, especially for small objects. Similar findings are also reported in~\cite{li2022delving}.

\begin{table}[htpb]
\begin{center}
\resizebox{1.0\linewidth}{!}{
\begin{tabular}{|l|c|c|}
\hline
Models & AP  VEHICLE ($\%$) & AP  HUMAN ($\%$) \\
\hline\hline
Single-camera baseline & 77.78 & 12.86 \\
Ours & 83.48 & 16.60 \\
\hline
\end{tabular}}
\end{center}
\caption{Comparing the proposed model (Ours) with a single-camera baseline at the IoU=$0.1$ threshold.}\label{tab:comparison_full_training}
\end{table}

\subsection{Training with Larger Data}
We experiment our model with $\times 20$ more data, and compare it with a single-camera baseline model, which runs detection on each monocular camera separately, and aggregates results from all cameras as the final multi-camera detection outputs (with non-maximum-suppression). The baseline model follows a network architecture similar to FCOS3D \cite{wang2021fcos3d}, which regresses cuboid parameters directly from 2D images. The baseline and our proposed models use the same pre-trained image backbone. Tab.~\ref{tab:comparison_full_training} shows the inference results on the same test subset in Sec.~\ref{subsec:main_results}. Our model (detr3d + feat, loc, query priors) outperforms the baseline model by $5.70\%$ and $3.74\%$ AP for VEHICLE and HUMAN classes, respectively. Besides, larger training data brings approx. $1.5\%$ performance gain, when comparing results from the small training data shown in Tab.~\ref{tab:comparison}. This marginal AP improvement suggests that the 2D priors from the image backbone might compensate the benefits from large dataset, saving the training cost.

\subsection{Training convergence}
\begin{figure}[!tpb]
\centering
\begin{minipage}{1\linewidth}
	\centering
	\includegraphics[width=1.0\textwidth]{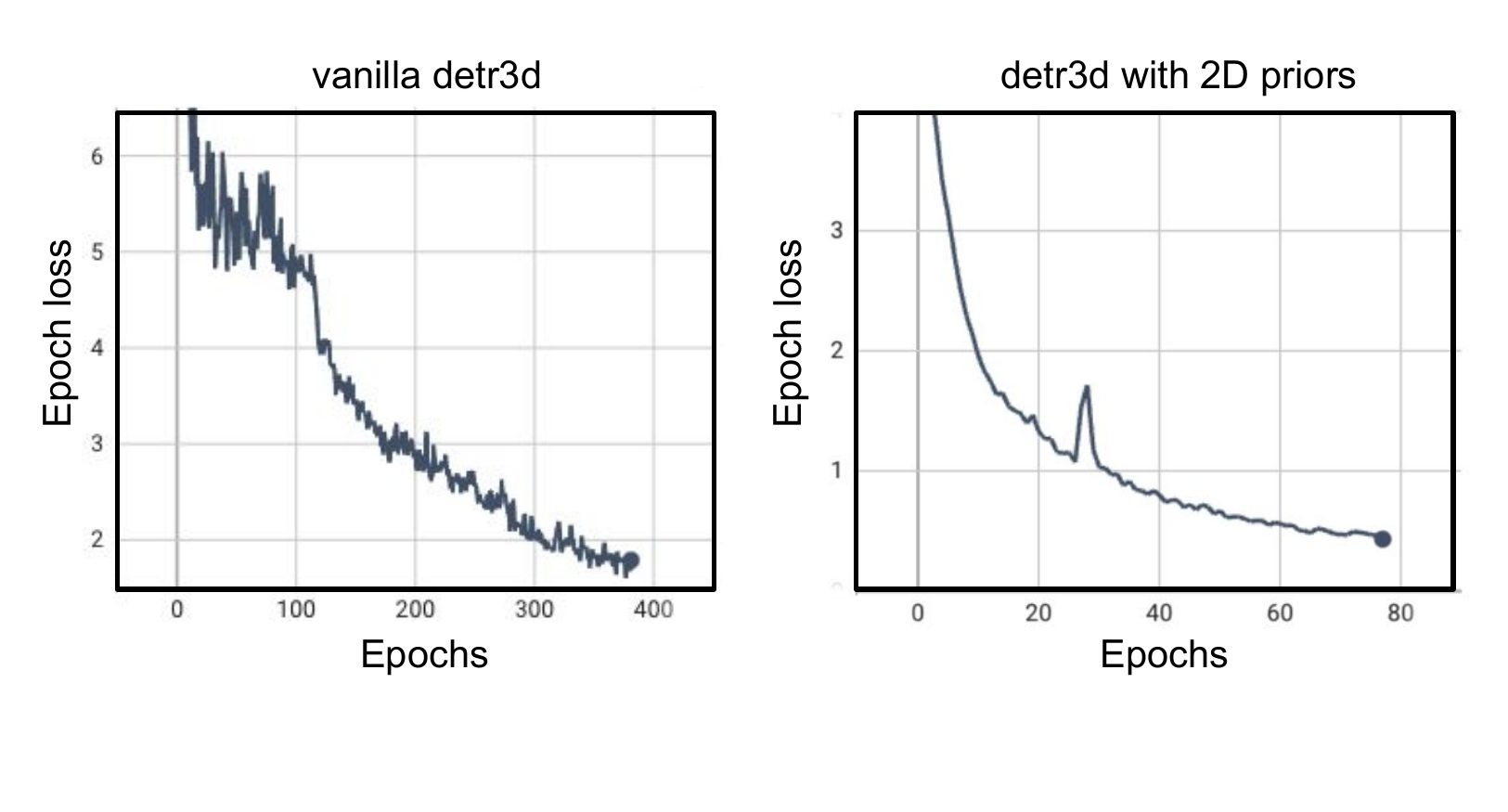}
	\caption{The learning curves for overfitting a small dataset.}
 \label{fig:learning_curve}
\end{minipage}
\end{figure}

We show the learning curves in Fig.~\ref{fig:learning_curve}, by overfitting a small dataset with approx. 300 data frames. Compared to the vanilla detr3d model, the model with 2D priors reaches the same epoch loss with much fewer epochs, implying the benefits of 2D priors for faster training convergence.
 
%-------------------------------------------------------------------------
\section{Summary} \label{sec:summary}
%-------------------------------------------------------------------------
Transformer-based methods advance the recent development of mulit-camera 3D detection. The vanilla transformer architecture randomly initializes queries, without considering the heterogeneity of inputs from different frame. We argue that this approach is sub-optimal in query generation. In this regard, we propose to leverage multiple predictions from an image backbone network as 2D priors to improve the transformer part of the network, including 2D detections, semantic maps, and depth maps. The method works by augmenting image feature maps with 2D priors, sampling query locations via ray-casting along 2D box centroids, as well as initialising query features with object-level image features. Experiments results show that 2D priors can be used to largely improve the detection accuracy in terms of average precision, and to accelerate the model convergence. In the future, we intend to add more 2D priors, such as scene flow and instance masks, and extend the framework into a multi-modal fusion setting (e.g. combining cameras, lidars, and radars)~\cite{feng2020leveraging, drews2022deepfusion, chen2022futr3d}.

%-------------------------------------------------------------------------
\section*{Acknowledgement}
The authors would like to thank the full detection team at Argo AI for the technical discussions and the ML infra support. Special thanks to Jan Martin and Ahsan Iqbal for making this publication possible.
%-------------------------------------------------------------------------

% references section
\bibliographystyle{IEEEtran}
\bibliography{bibliography}

\end{document}